\documentclass[10pt,twocolumn,letterpaper]{article}

\usepackage{cvpr}
\usepackage{times}
\usepackage{epsfig}
\usepackage{graphicx}
\usepackage{amsmath}
\usepackage{amssymb}

\usepackage[breaklinks=true,bookmarks=false]{hyperref}

\cvprfinalcopy 

\def\cvprPaperID{****} 

\begin{document}

\title{Exploring Low-light Object Detection Techniques}

\author{Winston Chen\\
{\tt\small wchen376@wisc.edu}
\and
Tejas Shah\\
{\tt\small tyshah2@wisc.edu}
}

\maketitle

\begin{abstract}
Images acquired by computer vision systems under low light conditions have multiple characteristics like high noise, lousy illumination, reflectance, and bad contrast, which make object detection tasks difficult. Much work has been done to enhance images using various pixel manipulation techniques, as well as deep neural networks - some focused on improving the illumination, while some on reducing the noise. Similarly, considerable research has been done in object detection neural network models. In our work, we break down the problem into two phases: 1) First, we explore which image enhancement algorithm is more suited for object detection tasks, where accurate feature retrieval is more important than good image quality. Specifically, we look at basic histogram equalization techniques and unpaired image translation techniques. 2) In the second phase, we explore different object detection models that can be applied to the enhanced image. We conclude by comparing all results, calculating mean average precisions (mAP), and giving some directions for future work.
\end{abstract}

\section{Introduction}
Lack of vision in low-light environments affects a person's ability to perform tasks. Computer vision systems that can help in such situations, especially for essential applications like visual surveillance or autonomous car driving, are highly valuable. However, factors like noise, glare, lousy illumination, low contrast, shadows, and reflectance make object detection in such conditions difficult.\\
\indent Much of the work in image enhancement is focused on subject beautification, and improving the aesthetic quality of the image, rather than feature retrieval. Some research works try to address the issue using improved hardware, like cameras equipped with infrared sensors and thermal images. These methods, however, are expensive, and the photos do not look realistic. Another approach is to use a suite of image enhancement techniques to improve image quality.\\
\indent Our work takes a holistic approach by comparing image enhancement techniques with object detection algorithms and analyzing results on different enhanced versions of images.

\section{Related Work}
Image enhancement techniques can be broadly classified into two categories - those based on pixel manipulation, and those on features (typically image translations using convolutional neural networks (CNNs)). Applying uniform enhancement techniques across an entire image does not always ensure good results due to the non-uniformity of luminance in a scene. There might be multiple sources of light in the picture, and accordingly, we need to adjust our enhancement function for good feature retrieval, which makes this a non-trivial task. Recent research works to address this issue include Gladnet\cite{wang2018gladnet}, which is a low-light enhancement network with global awareness, and \cite{Gaussian}, which proposes a Gaussian Process regression to construct a distribution of localized feature enhancement functions with the support of CNNs.\\
\indent For object detection, modern deep-learning based object detectors can be broadly classified as two-stage detectors (Region-based models like R-CNN, Fast R-CNN, Faster R-CNN) or single-stage detectors (like YOLO, SSD). For our work, we explore with variants of Faster R-CNN \cite{DBLP:journals/corr/RenHG015} and SSD \cite{DBLP:journals/corr/LiuAESR15}.\\
\indent One of the major challenges we face in low-light object detection is collecting paired datasets (low-light image and corresponding regular-light image) with object annotations. Most of the work done in this domain uses synthetically generated image pairs (using image editing software like Photoshop, or Python OpenCV packages). Another option is to manually take corresponding low-light and regular light photographs, and annotate the objects, which is pretty time-consuming. Hence, we relied on unpaired image-to-image translation using CycleGAN \cite{CycleGAN2017} for this work.
\section{Methodology}
\subsection{Dataset}
ExDark \cite{Exdark} is a dataset of exclusively dark images for low-light object detection. It collects images comprising 12 object categories under ten different low light conditions - "low contrast", "ambient", "high-contrast object", "single light-source", "weak luminance", "strong light-source", "screen/shadow", "window/twilight", in both indoor and outdoor scenarios. This dataset covers a wide range of poor light conditions and common subjects ("boat", "bicycle", "chair", "bus", "cat", "car", "table", "bottle", "dog", "motorbike", "people" and "cup"). 
\indent MS-COCO dataset \cite{lin2014microsoft} is a well-known dataset containing nearly 118K images of 100 subjects. It is one of the golden standards to evaluate object detection models, so several popular models are pre-trained on the MS-COCO dataset and made open to public for use \cite{tf_model_zoo}.
\subsection{Models}
We select Faster-RCNN with ResNet-50 as backbone and SSD with ResNet-50 FPN as backbone for our object detection tasks. These models were taken from Tensorflow-Model-Zoo, pre-trained on the MS-COCO dataset. \\
\indent We use two cycleGAN \cite{CycleGAN2017} models, with different generators for unpaired image-to-image translation. The first one uses a 9 residual block based generator, consisting of a series of Convolution-Instance\_Normalization-ReLu layers as an encoder to extract features. In this generator model, residual blocks combine nearby features and decide feature modifications in the transformer stage, and several deconvolution layers expand the low-level features to build the output image in the target domain. The second CycleGAn model uses U-net as a generator, containing an encoder with max-pooling to downsample images and extract features, a decoder to upsample and reconstruct images, and residual pathways connecting the same level layers for more extensive capability to shape the output. Both CYcleGAN models use 70x70 PatchGAN \cite{pathcGAN} architecture for discriminator, which outputs a probability matrix determining whether a particular patch of image is real or fake.

\subsection{Roadmap}
Firstly, we need to enhance images to brighten regions with low luminance. Second, we deploy models on the adjusted images to detect objects. Additionally, original dark images are evaluated by our selected models to give baseline numbers, and identify the influence of enhancement techniques. 
\section{Implementation}
\subsection{Brightness histogram normalization}
\begin{figure}[h]
\centering
\includegraphics[width=0.4\textwidth]{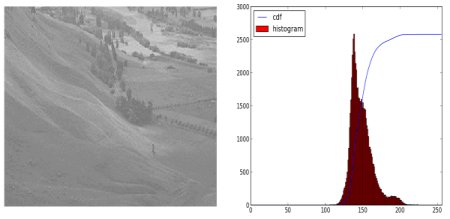}
\caption{An image and histogram before HE func.}
\label{he_1}
\end{figure}

\begin{figure}[h]
\centering
\includegraphics[width=0.4\textwidth]{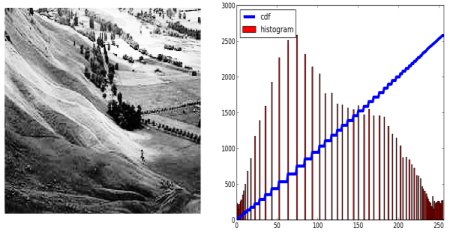}
\caption{An image and histogram after HE func.}
\label{he_2}
\end{figure}

This technique is an intuitive idea, stretch the brightness distribution. We first convert the input image format from RGB to YUV, so that we can directly manipulate brightness. Then, we apply Histogram Equalization(HE) function of OpenCV\cite{opencv_python} to extend the brightness range to either end, as shown in Figure \ref{he_1} and Figure \ref{he_2}.

\subsection{Image-to-image translation}
Here we use Cycle-GAN \cite{CycleGAN2017} to perform the image-to-image translation of unpaired datasets. Random images are selected from the MS-COCO dataset of the same categories as the ExDark dataset. The sample populations of two datasets stay equivalent, to meet a diversity-speed balance. Two generator architectures are applied, ResNet\_9 and Unet\_256. The cycleGAN models learned to transfer images between two styles: regular luminance and low luminance. For 200 epochs, the ResNet model took around seven days to train using a GTX 1080 GPU with batchsize=3, and the Unet model took two days to train using the same GPU with batchsize=18. Under the same 8GB memory of GPU, the ResNet model occupied more space, so the batch size is smaller. The transformation result is shown in Figure \ref{bus}\\

\subsection{Object detection}
After getting better images, we perform object detection on the four datasets: original ExDark, basic enhancement, cycleGAN-ResNet, and CycleGAn-Unet. Two pre-trained models from Tensorflow Object-detection Zoo are utilized, named as "faster\_rcnn\_resnet50\_coco" and "ssd\_resnet\_50\_fpn\_coco". These two models generate bounding boxes and confidence for each image. The annotations from the ExDark repository are used as ground truth to evaluate the performance of various datasets. 
\section{Results and discussion}
\subsection{Image enhancement}
\begin{figure*}[p]
\centering
  \includegraphics[width=0.7\textwidth]{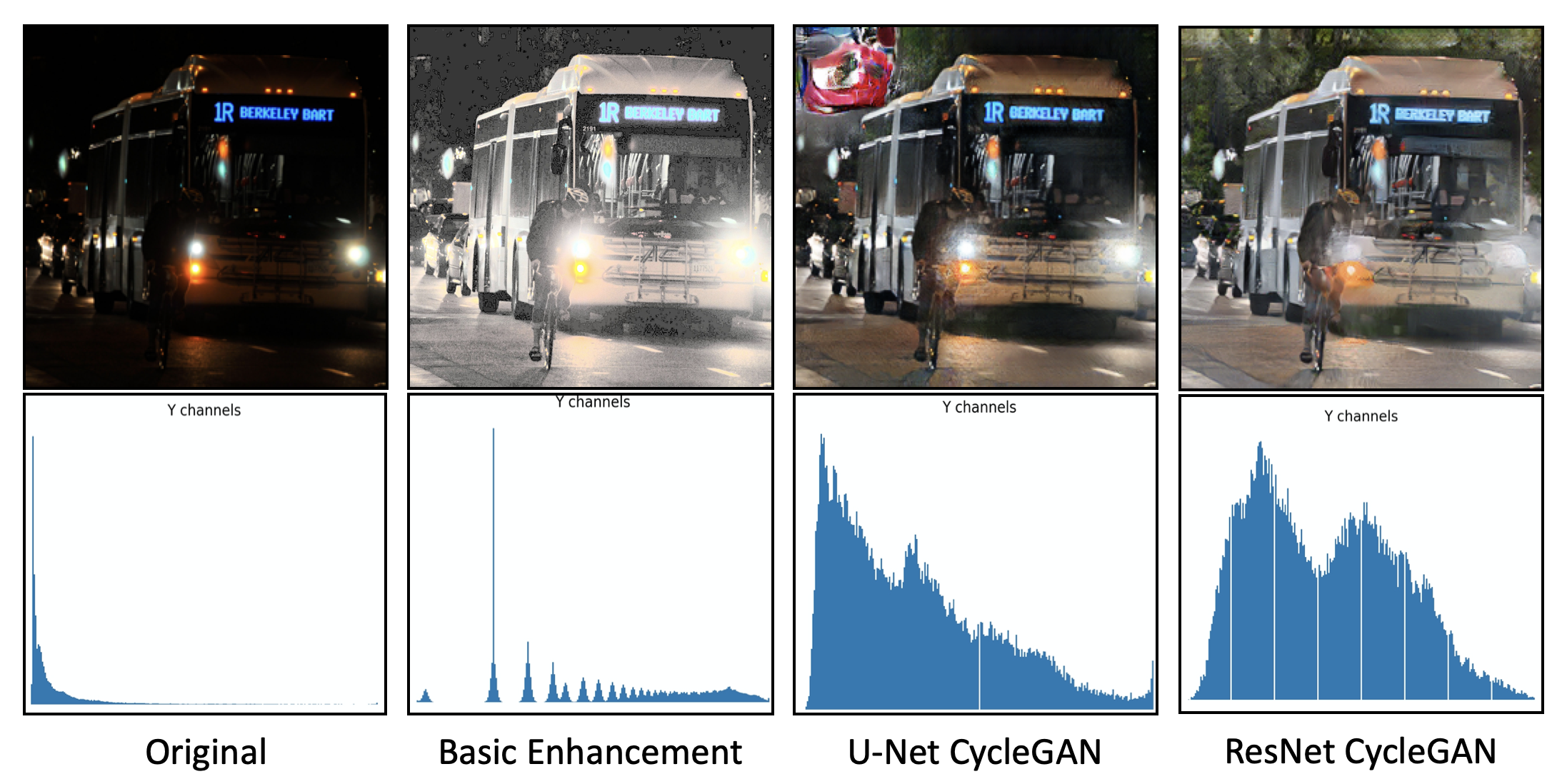}
\caption{Images and corresponding pixel histogram of original images and three methods of enhancement.}
\label{bus}
\end{figure*}

As shown in Figure \ref{bus}, the original image looks dark, and it's hard to identify the bicycle before the bus. After the basic enhancement (Histogram Equalization), although the brightness is improved, the pixel distribution is discrete and concentrates on several specific values. The pixel distribution over the histogram is more uniform across the entire range after cycleGAN enhancement. However, cycleGAN-Unet generates some artifacts (as is visible in the upper left corner of the image), and cycleGAN-ResNet leads to better image quality, and histogram uniformity. Also, the bus front is reconstructed better in the ResNet-CycleGAN model, as compared to the U-Net CycleGAN model. Please note that since we are using unpaired datasets, no ground-truth of this image can be evaluated as standard to show what is the correct pixel distribution.
\subsection{Object detection}
\begin{figure*}[p]
\centering
   \includegraphics[width=0.7\textwidth]{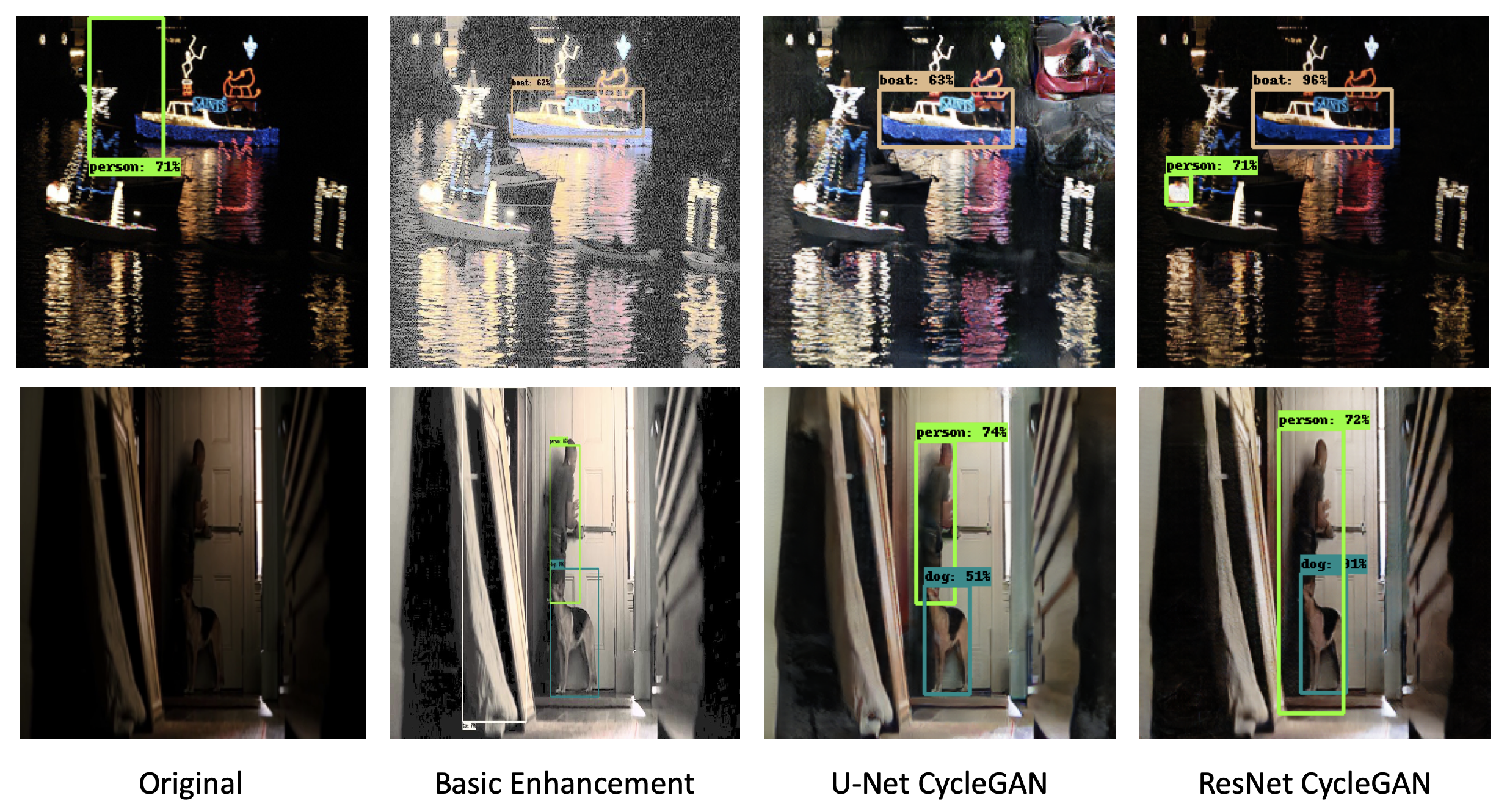}
\caption{Object detection results based on different datasets.}
\label{bbox}
\end{figure*}

We consider two samples, as shown in Figure \ref{bbox}, to analyze the influence of the image enhancement techniques on object detection accuracy. The first row in Figure \ref{bbox} aims to detect boats over the water. The bounding box of the original image wrongly recognizes a region as a "person." After basic enhancement, the object detection model is able to identify the boat, but with low confidence. After the enhancement of U-Net based cycleGAN, the boat can be detected accurately. The ResNet based CycleGAN model enables detection of the boat, as well as a person on one of the boats. For the second row of images in \ref{bbox}, a person and a dog are located in the image. The original image means nothing to the object detection models. Although the model can detect the person and dog from a basic enhancement image, it misrecognized an object as "tie" (in front of the person and dog). The two cycleGAN enhanced images perform well, and the ResNet one leads to a more accurate bounding box.
\subsection{Mean Average Precision (mAP)}
\begin{figure*}[p]
\centering
   \includegraphics[width=0.7\textwidth]{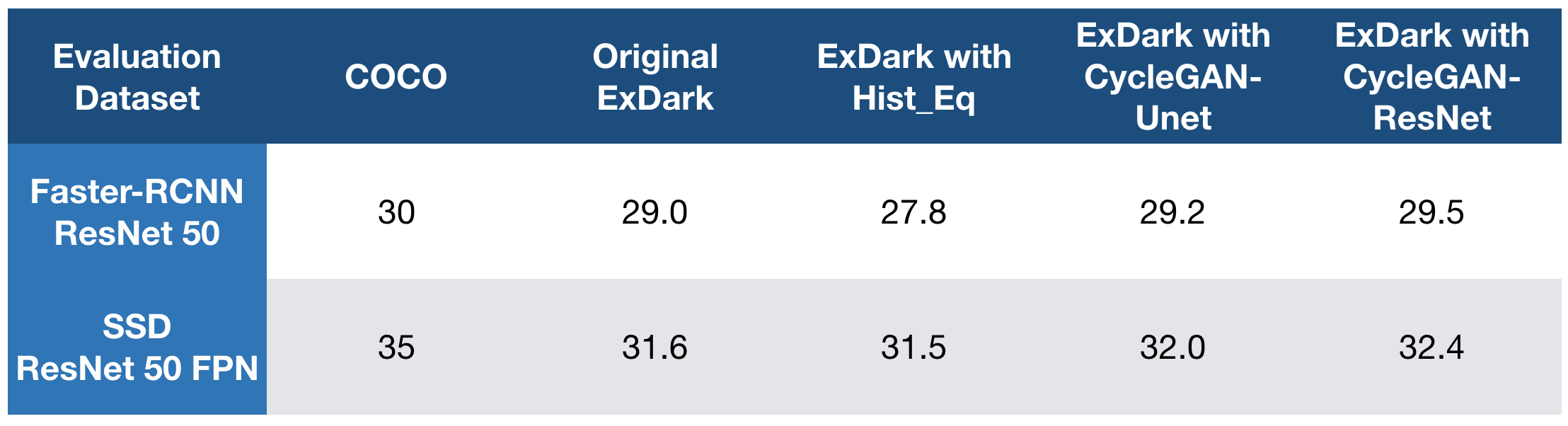}
\caption{Object detection results of mAP based on different datasets.}
\label{mAP}
\end{figure*}

In Figure \ref{mAP}, we compare the overall performance mAP. The first column comes from Tensorflow Model Zoo, evaluated by "COCO 14 minival set"\cite{tf_model_zoo}. If we take the "Orginal ExDark" as a baseline, it becomes obvious that the histogram equalization technique results in accuracy loss for object detection. The two CycleGAN model enhancements improve mAP over the baseline model on the ExDark dataset to some extent. The ResNet version leads to more considerable improvement, which can be attributed to its better image quality generation, larger model size, and hence longer training time. We want to note here, however, that the resultant mAPs are still lesser than the object detection results on the MS-COCO dataset.

\section{Conclusion}
It can be seen from the object detection results that basic image enhancement techniques like modifying the brightness using histogram equalization harm object detection accuracy. This is so because such image enhancements are not adaptive, and apply uniformly across the image, to all images in the dataset. Adaptive enhancement techniques relying on image features (CycleGAN outputs) give better results, which is ascertained by the mAP numbers. Also, the ResNet-based CycleGAN performs better than the U-Net based CycleGAN. Still, it takes much longer to train -- hence, one can decide to use a specific model depending on the accuracy requirements of the application.
\section{Future Work}
In the future, we would like to explore image enhancement techniques using pixel enhancement functions that adapt to the non-uniform scene luminance, like the one mentioned in \cite{Gaussian}. This might give better control over features getting generated in the output image. Also, we could try to combine the image enhancement and object detection steps into a single pipeline, like by integrating an image enhancement model into the backbone network of Faster-RCNN. 

\section{Individual Contributions}
\noindent Each group member's work is listed below:
\begin{enumerate}
    \item Winston Chen - Basic image enhancements, object detection experiments, mAP calculation.
    \item Tejas Shah - CycleGAN training and experimentation, dataset collection.
\end{enumerate}
Overall, both team members contributed equally towards the project, either directly, or through code reviews. The source code for the project is available at \cite{CS_838}.


\clearpage
\bibliographystyle{ieee_fullname}
\bibliography{egbib}

\end{document}